\documentclass[10pt]{article}
\usepackage{amsmath,amsthm,amsfonts,amssymb,latexsym,graphicx}
\usepackage[noadjust]{cite}
\usepackage{algorithm}
\usepackage[noend]{algpseudocode}
\algrenewcommand\algorithmicthen{\relax}
\algrenewcommand\algorithmicdo{\relax}
\usepackage[pdfpagemode=UseNone,pdfstartview=FitH]{hyperref}
\newcommand{\Extra}[1]{}

\newcommand{\OCMXXIV}{Vovk:2021}
\newcommand{\OCMXXXI}{Vovk:arXiv2012}
\newcommand{\OCMXXXII}{Vovk/etal:arXiv2102}
\newcommand{\OCMXXXIII}{Vovk:arXiv2104}

\allowdisplaybreaks[4]

\theoremstyle{definition}

\newtheorem{remark}{Remark}

\newcommand*{\dd}{\,\mathrm{d}}
\newcommand*{\ddd}{\mathrm{d}}

\newlength{\IndentI}
\newlength{\IndentII}
\newlength{\IndentIII}
\newlength{\IndentIV}
\newlength{\WidthI}
\newlength{\WidthII}
\newlength{\WidthIII}
\newlength{\WidthIV}
\title{Enhancement of prediction algorithms by betting}
\author{Vladimir Vovk}

\begin{document}
\maketitle

\begin{abstract}
  This note proposes a procedure for enhancing the quality of probabilistic prediction algorithms
  via betting against their predictions.
  It is inspired by the success of the conformal test martingales that have been developed recently.

   The version of this note at \url{http://alrw.net} (Working Paper 34)
   is updated most often.
\end{abstract}

\section{Introduction}

This note is inspired by the power of betting martingales used in conformal prediction;
e.g., the conformal test martingales developed in \cite{\OCMXXXI,\OCMXXXII,\OCMXXXIII}
are much more successful as compared with the older ones \cite{Vovk/etal:2005book,\OCMXXIV}.
However, the method that it proposes is independent of conformal prediction
and, in particular, does not depend on the IID assumption and can be applied, e.g., to time series.

L\'evy's \cite[Section 39]{Levy:1937} probability integral transform translates testing a prediction algorithm
that outputs probabilistic predictions (a probability forecasting system)
into testing independent uniformly distributed random variables in $[0,1]$.
The numerous betting martingales developed in conformal testing often allow us
to multiply the initial capital manyfold on real-world datasets
(see, e.g., the results for the USPS dataset in \cite{\OCMXXXI}).
Such betting martingales translate into test martingales that gamble successfully
against the original probability forecasting system.
However, since a test martingale is essentially the same thing as the likelihood ratio
with the original probability forecasting system in the denominator,
a successful test martingale provides us with a new (\emph{enhanced}) probability forecasting system
(the one in the numerator)
that outperforms the original probability forecasting system.

The idea of this note is to use testing procedures
(namely, betting martingales, such as the Sleepy Jumper and Mean Jumper)
for developing better prediction algorithms.
In this respect it is reminiscent of the method of defensive forecasting \cite[Chapter 12]{Shafer/Vovk:2019},
which starts from a test martingale (more generally, a strategy for Sceptic) and then develops a prediction algorithm
that prevents the test martingale (more generally, Sceptic's capital) from growing.
An advantage of our current procedure is that in typical cases it is computationally more efficient
(in particular, it never requires finding fixed points or solving equations,
as in defensive forecasting).

We will start in Section~\ref{sec:PIT} from discussing L\'evy's probability integral transform,
which generates IID random variables distributed uniformly in $[0,1]$
(the analogue of p-values in conformal prediction).
The topic of Section~\ref{sec:testing} is online testing of the output of the probability integral transform for uniformity.
Section~\ref{sec:enhancement} combines results of the previous two sections
for the purpose of enhancing prediction algorithms.
A toy simulation study is described in \ref{sec:simulation},
but the enhancement procedure of Section~\ref{sec:enhancement} is general and widely applicable;
this will be further discussed in Section~\ref{sec:conclusion}.

\section{Probability integral transform}
\label{sec:PIT}

The key fact that makes conformal testing possible is that conformal prediction outputs p-values
that are independent and distributed uniformly in $[0,1]$.
Without assuming that the observations are IID,
we have a similar phenomenon for the probability integral transform:
if a probability forecasting system outputs probabilistic forecasts
with distribution functions $F_1,F_2,\dots$ and $y_1,y_2,\dots$ are the corresponding observations,
the values $F_1(y_1),F_2(y_2),\dots$ are independent and distributed uniformly in $[0,1]$.
(To avoid complications, let us assume that all $F_n$ are continuous.)

The uniformity of the probability integral transform
was used by L\'evy \cite[Section~39]{Levy:1937} as the foundation
of his theory of denumerable probabilities
(which allowed him to avoid using the then recent axiomatic foundation
suggested by Kolmogorov in his \emph{Grundbegriffe} \cite{Kolmogorov:1933}).
Modern papers, including \cite{Dawid/Vovk:1999}, usually refer to Rosenblatt \cite{Rosenblatt:1952},
who disentangled L\'evy's argument from his concern with the foundations of probability;
Rosenblatt, however, refers to L\'evy's 1937 book \cite{Levy:1937} in his paper.

The probability integral transform can be used for testing the underlying probability forecasting system
considered as the data-generating distribution.
See, e.g., \cite[Sections 3.8 and 4.7]{Dawid/Vovk:1999}.

\section{Testing probability forecasting systems}
\label{sec:testing}

\begin{algorithm}[bt]
  \caption{Jumper betting martingale ($(u_1,u_2,\dots)\mapsto(S_1,S_2,\dots)$)}
  \label{alg:SJ}
  \begin{algorithmic}[1]
    \State $C_{-1}:=C_0:=C_1:=1/3$
    \State $C:=1$
    \For{$n=1,2,\dots$:}
      \For{$\epsilon\in\{-1,0,1\}$:}
        $C_{\epsilon} := (1-J)C_{\epsilon} + (J/3)C$
      \EndFor
      \For{$\epsilon\in\{-1,0,1\}$:}
        $C_{\epsilon} := C_{\epsilon} f_{\epsilon}(u_n)$
      \EndFor
      \State $S_n := C := C_{-1}+C_0+C_1$
    \EndFor
  \end{algorithmic}
\end{algorithm}

To transform the probability integral transforms $u_1,u_2,\dots$ into a test martingale
we use, as in \cite{\OCMXXXI,\OCMXXXII,\OCMXXXIII},
the \emph{Simple Jumper} betting martingale given as Algorithm~\ref{alg:SJ},
where
\begin{equation}\label{eq:f}
  f_{\epsilon}(u)
  :=
  1 + \epsilon(u-0.5).
\end{equation}
In the next section we set $J:=0.01$, as in \cite{\OCMXXXII}.
For the intuition behind Simple Jumper,
see \cite{\OCMXXXII}
(and the more complicated Sleepy Jumper is described in detail in \cite[Section~7.1]{Vovk/etal:2005book}).

A safer option than the Simple Jumper is the Mean Jumper betting martingale \cite{\OCMXXXI},
which is defined to be the average of Simple Jumpers
over a finite set $\mathcal{J}$ of $J$ including $J=1$ (such as $J\in\mathcal{J}:=\{10^{-3},10^{-2},10^{-1},1\}$).
The inclusion of $J=1$ is convenient since the corresponding Simple Jumper is identical 1,
and so the Mean Jumper never drops in value below $1/\left|\mathcal{J}\right|$.

\section{The enhancement procedure}
\label{sec:enhancement}

Given a prediction algorithm $A$ and a betting martingale $S$,
our enhancement procedure produces the prediction algorithm $A'$
such that $S$ is the likelihood ratio process $\ddd A'/\ddd A$.

If the predictive distribution function output by $A$ is $F_n=F_n(y)$,
the corresponding predictive density is $f_n=f_n(y)=F'_n(y)$,
and the betting function output by $S$ is $b_n$,
the enhanced predictive density is $b_n(F_n)f_n$.
It integrates to 1 since $b_n(F_n)f_n=(B_n(F_n))'$,
where $B_n$ is the indefinite integral $B_n(v):=\int_0^v b_n$ of $b_n$, so that $B'_n=b_n$.
We can see that the distribution function for the enhanced algorithm is $B_n(F_n)$.

In this note we evaluate the performance of the original and enhanced algorithms
using the log loss function;
namely the loss of a predictive density $f$ for a realized outcome $y$ is $-\log f(x)$
(two pictures in the experimental Section~\ref{sec:simulation} will use base 10 logarithms).
It is a proper loss function,
meaning that the optimal expected loss is attained by the true predictive density
\cite[Section 4]{Gneiting/Raftery:2007}.

When $S$ is a Mean Jumper martingale,
the enhanced algorithm is guaranteed not to lose much as compared with the original algorithm
when the quality is measured by the log loss function:
namely, the cumulative log loss for $A'$ is at most the cumulative log loss for $A$ plus $\log\left|\mathcal{J}\right|$.

\section{A simulation study}
\label{sec:simulation}

We consider a dataset that consists of independent Gaussian observations:
the first 1000 are generated from $N(0,1)$, and another 1000 from $N(1,1)$.
Our base prediction algorithm does not know that there is a changepoint at time 1000
and always predicts $N(0,1)$.
The seed of the pseudo random number generator (in NumPy) is always 2021.

\begin{figure}
  \begin{center}
    \includegraphics[width=0.48\textwidth]{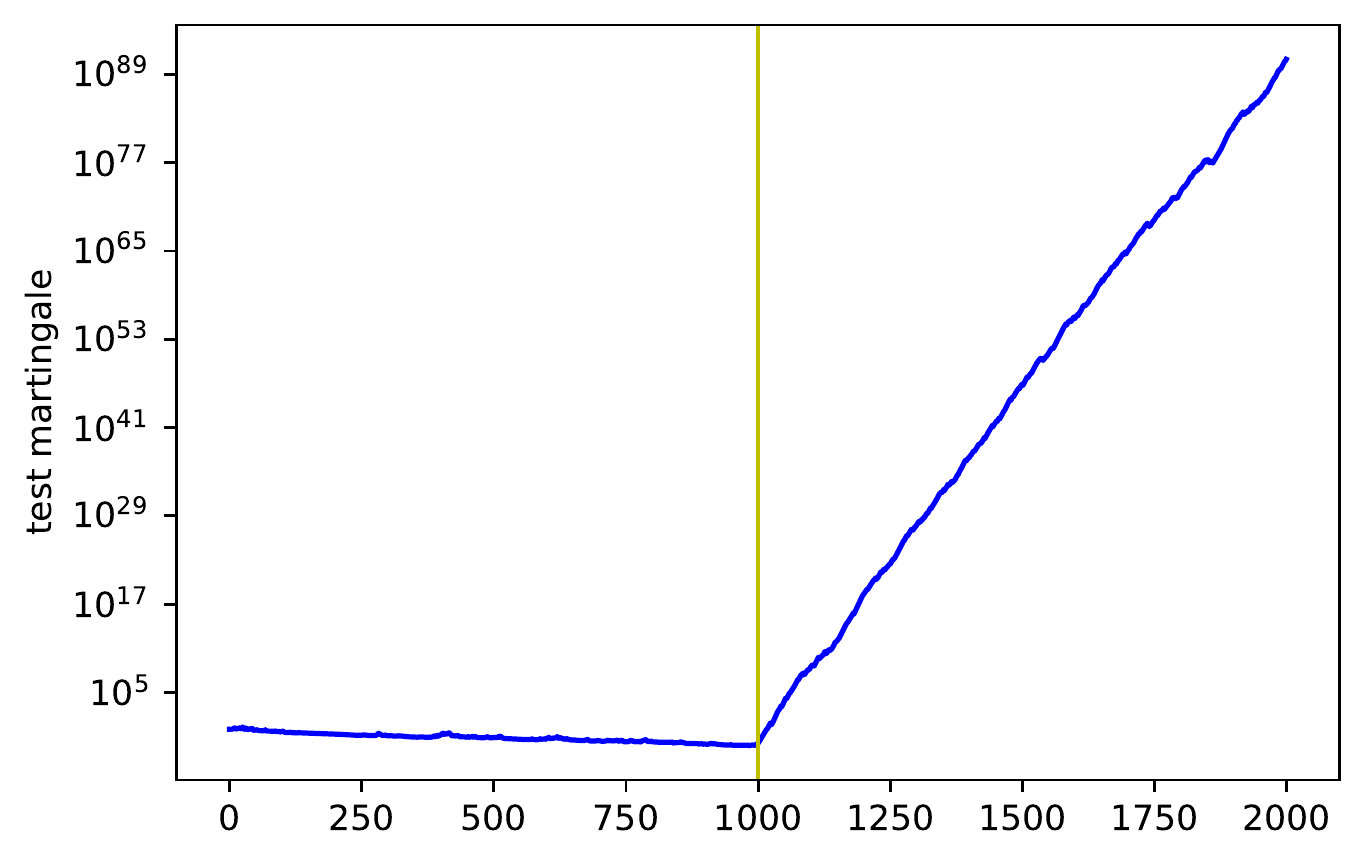}
    \includegraphics[width=0.48\textwidth]{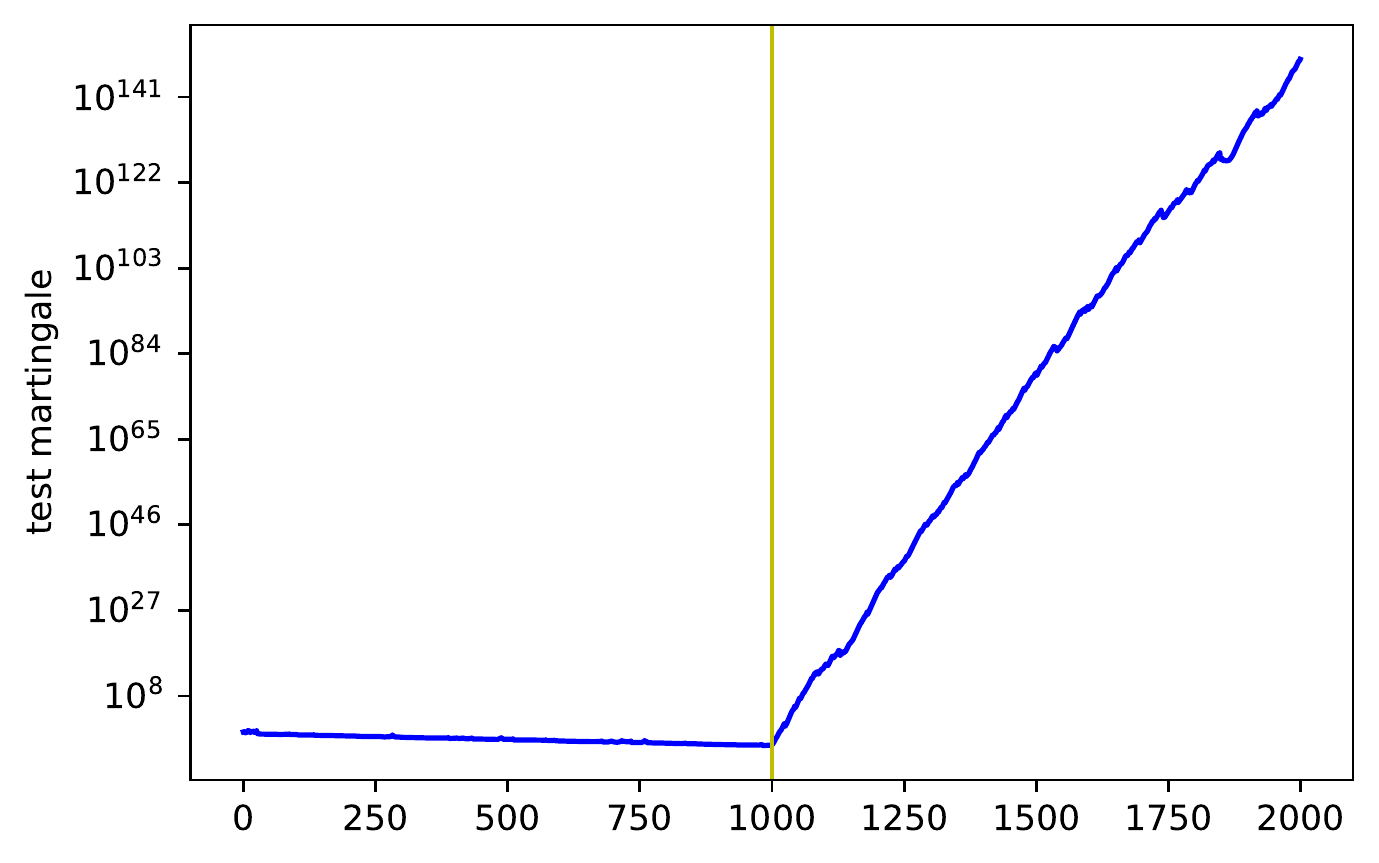}
  \end{center}
  \caption{The Simple Jumper test martingale.
    Left panel: the standard one ($\epsilon\in\{-1,0,1\}$).
    Right panel: $\epsilon\in\{-2,0,2\}$.}
  \label{fig:SJ}
\end{figure}

First we run the Simple Jumper martingale (Algorithm~\ref{alg:SJ} with $J=0.01$) on our dataset.
The left panel of Figure~\ref{fig:SJ} shows its trajectory;
it loses capital before the changepoint,
but quickly regains it afterwards.
Its final value is $1.100\times10^{91}$.

\begin{remark}
  The possibility of losing so much capital before the changepoint
  (the value of the Simple Jumper at the changepoint is $0.0114$)
  shows that using the Simple Jumper is risky.
  If we want to play safe, we can use the Mean Jumper instead of the Simple Jumper.
  As mentioned above, this will bound our loss to $\log\left|\mathcal{J}\right|$
  as compared with the original algorithm.
\end{remark}

\begin{figure}
  \begin{center}
    \includegraphics[width=0.48\textwidth]{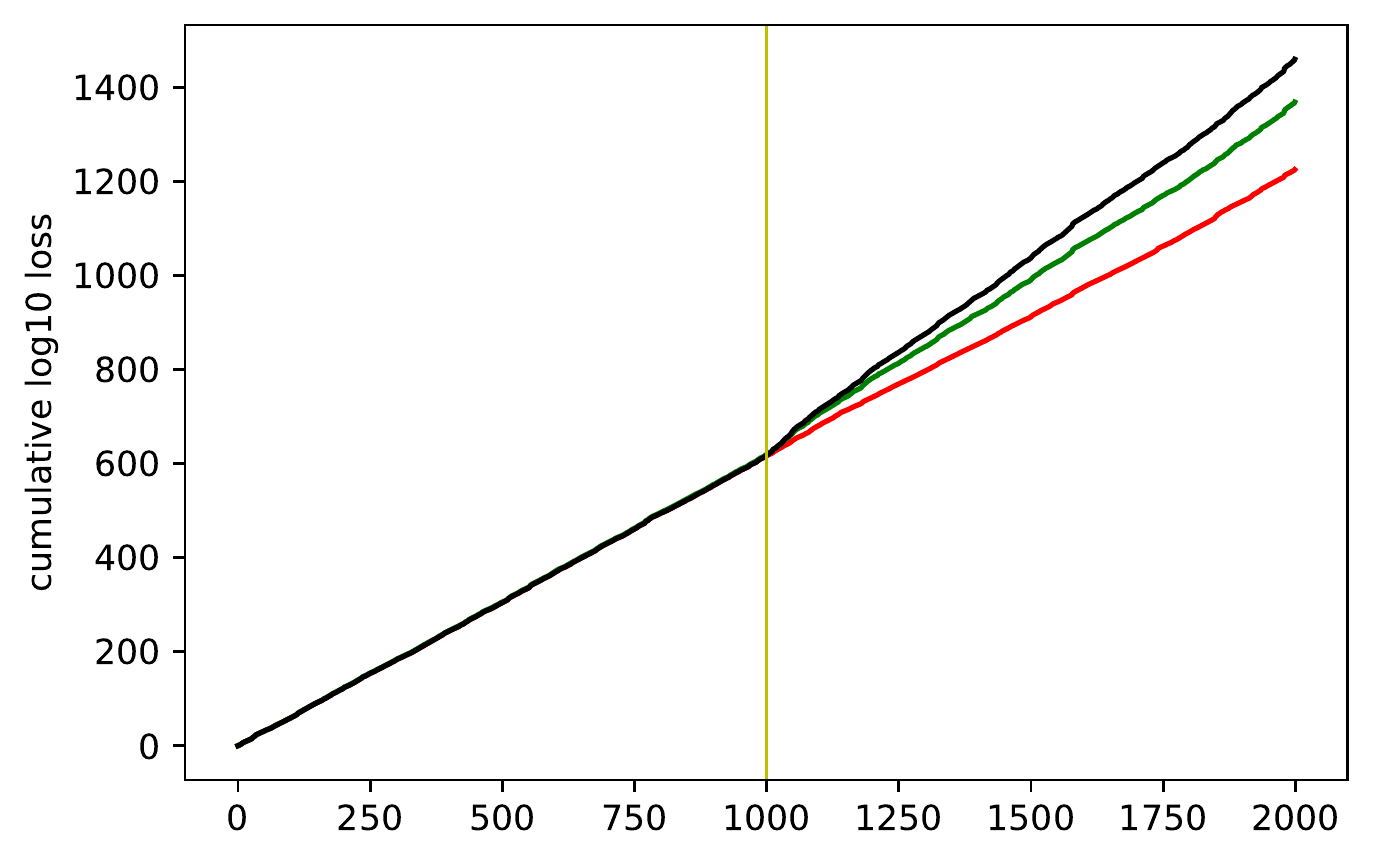}
    \includegraphics[width=0.48\textwidth]{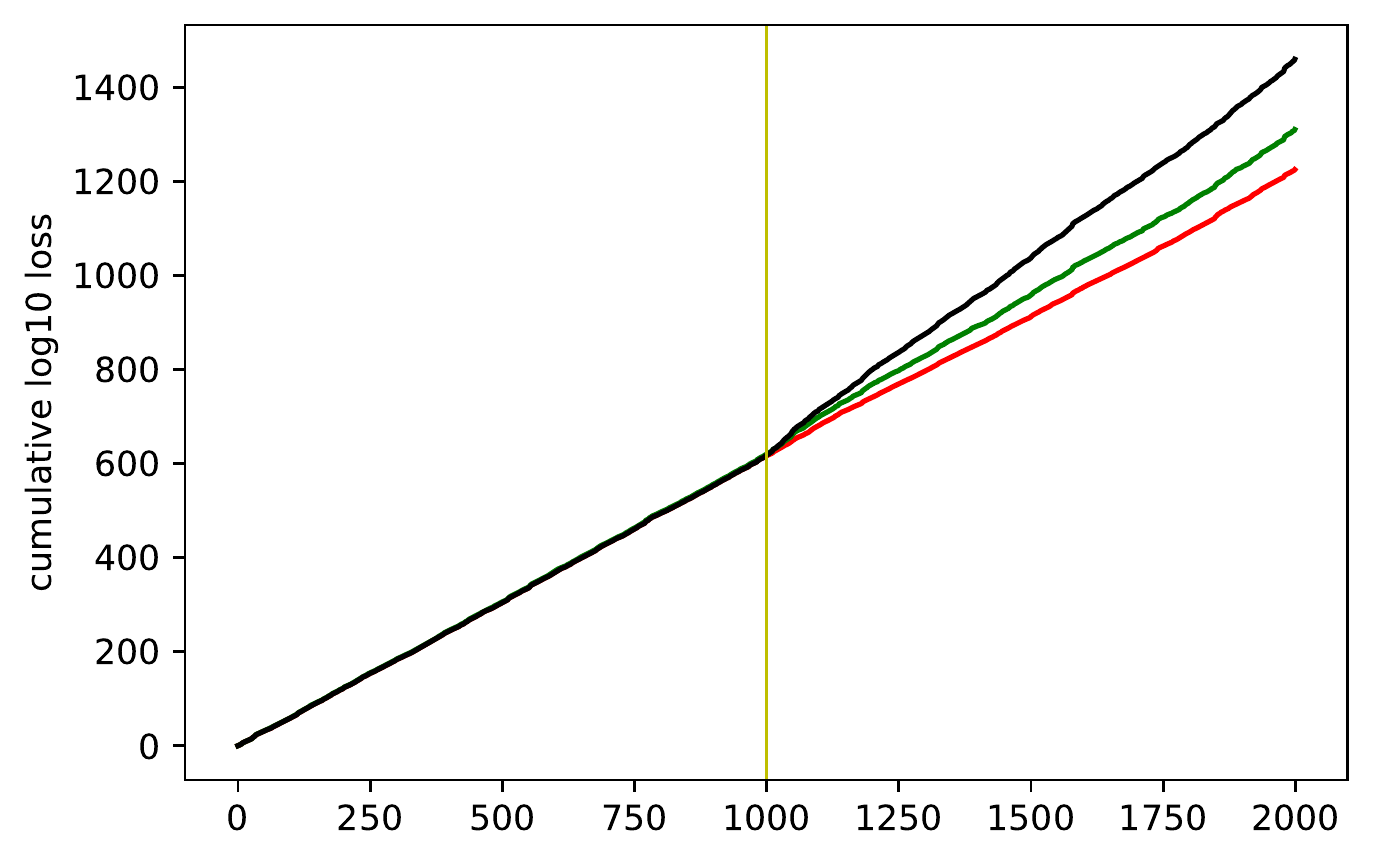}
  \end{center}
  \caption{The cumulative log losses of three prediction algorithms
    (to the left of the changepoint the three lines coincide or are visually indistinguishable).
    Left panel: $\epsilon\in\{-1,0,1\}$.
    Right panel: $\epsilon\in\{-2,0,2\}$.}
  \label{fig:log-loss}
\end{figure}

The cumulative log loss of the enhanced version of the base prediction algorithm
is shown as the green line in the left panel of Figure~\ref{fig:log-loss}.
The black line corresponds to the base algorithm,
and the red line to the impossible \emph{oracle algorithm},
which knows the truth and predicts with $N(0,1)$ before the changepoint and $N(1,1)$ afterwards.
According to Figure~\ref{fig:SJ} (left panel),
the difference between the final values of the black and green lines is about 91.

\begin{figure}
  \begin{center}
    \includegraphics[width=0.48\textwidth]{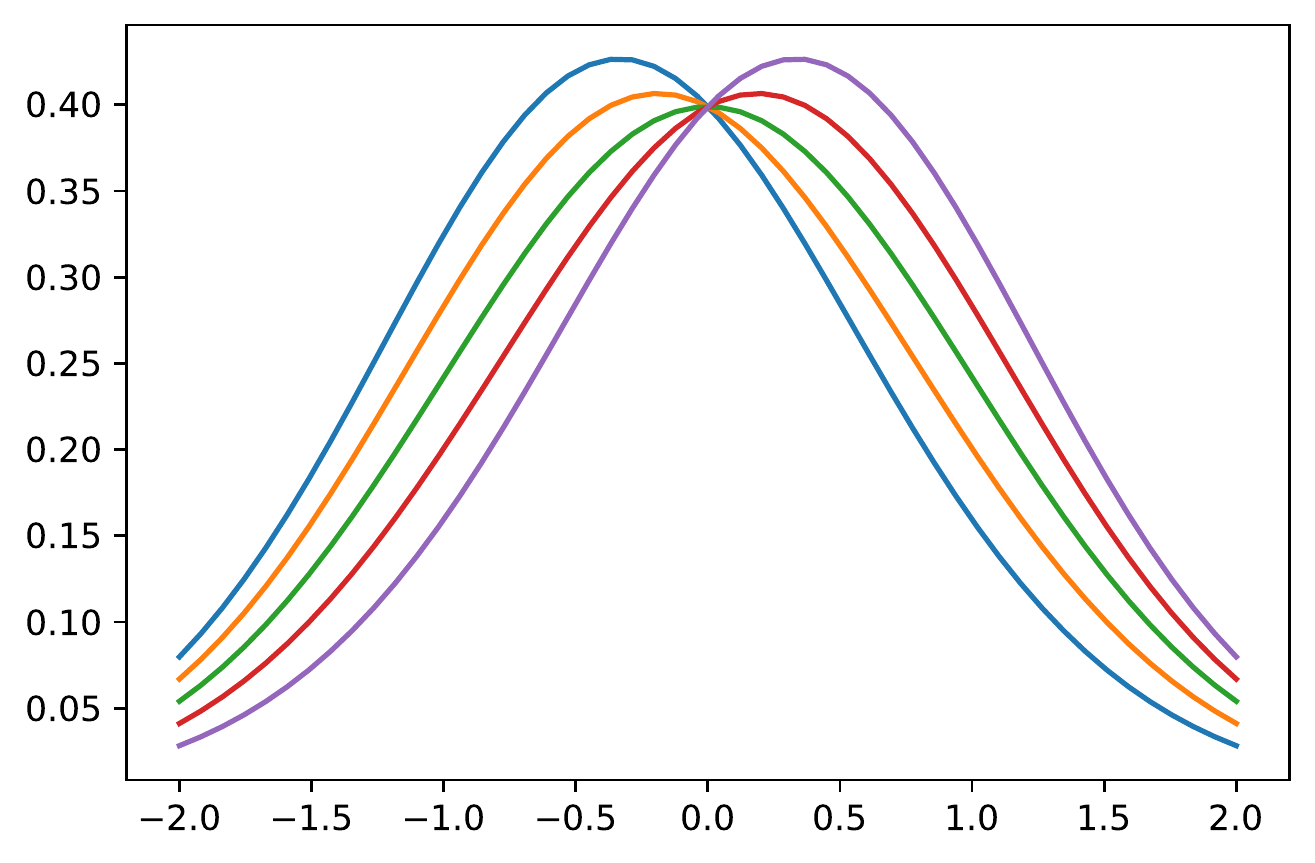}
    \includegraphics[width=0.48\textwidth]{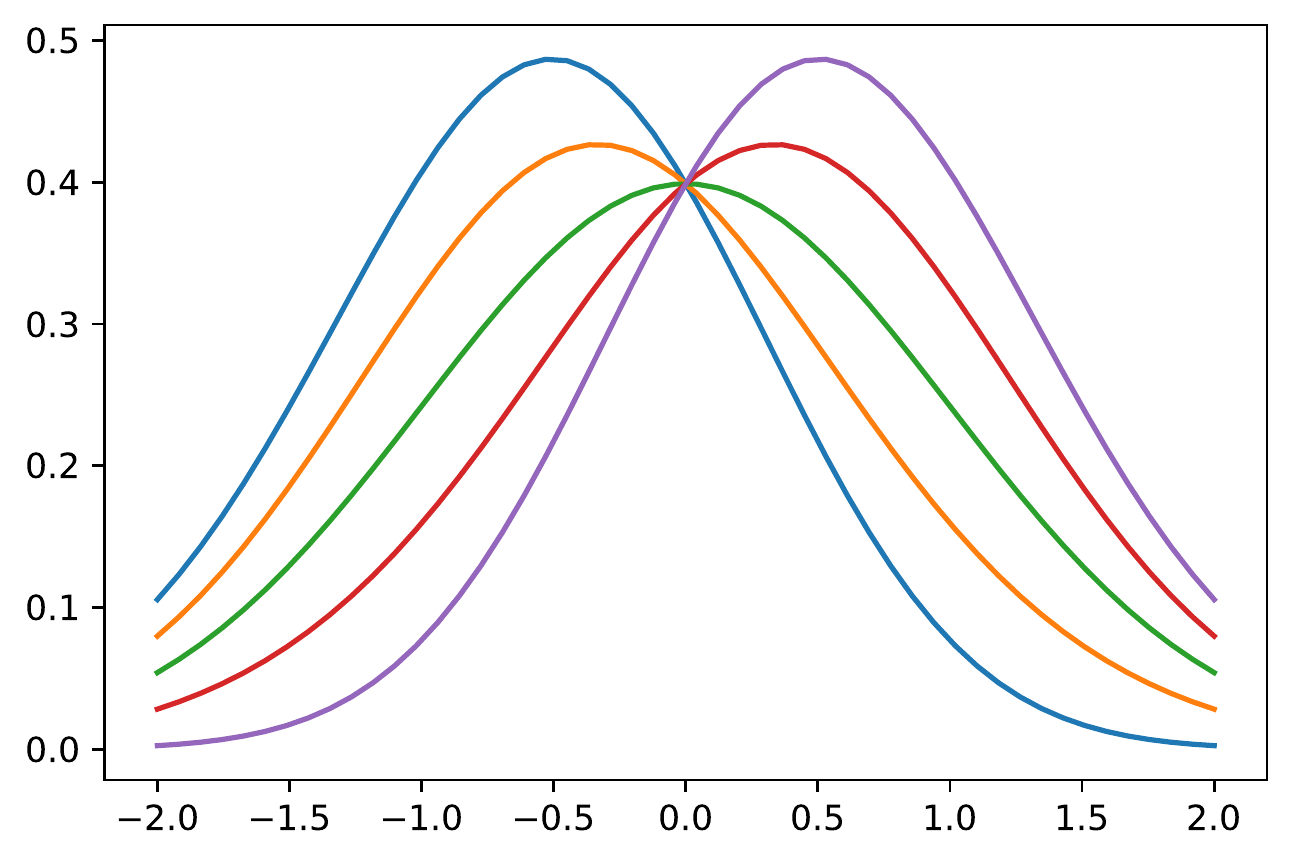}
  \end{center}
  \caption{The enhanced predictive distributions.
    Left panel: the range of $\epsilon$ is $\{-1,-0.5,0,0.5,1\}$.
    Right panel: $\epsilon\in\{-2,-1,0,1,2\}$.}
  \label{fig:enhanced}
\end{figure}

To understand better the mechanism of enhancement in this case,
notice that the Simple Jumper outputs betting functions $b$ of the form \eqref{eq:f},
where $\epsilon\in[-1,1]$ (usually $\epsilon\notin\{-1,0,1\}$).
The corresponding predictive distributions $b(F)f$
(where $f$ is the standard normal density and $F$ its distribution function)
are shown in the left panel of Figure~\ref{fig:enhanced} for five values of $\epsilon$.
We can see that our range of $\epsilon$, $\epsilon\in[-1,1]$, is not sufficiently ambitious
and does not allow us to approximate $N(1,1)$ well.

Replacing the range $\{-1,0,1\}$ for $\epsilon$ by $\{-2,0,2\}$,
we obtain the right panel of Figure~\ref{fig:enhanced}.
The right-most graph in that panel now looks closer to the density of $N(1,1)$.
We cannot extend the range of $\epsilon$ further without \eqref{eq:f} ceasing to be a calibrator.
(Of course, the calibrator does not have to be linear,
but let us stick to the simplest choices in this version of the paper.)

Using the range $\{-2,0,2\}$ for $\epsilon$ leads to the right panels
of Figures~\ref{fig:SJ} and~\ref{fig:log-loss}.
We can see that in the right panel of Figure~\ref{fig:log-loss}
the performance of the enhanced algorithm is much close to that of the oracle algorithm
than in the left panel.

\begin{figure}
  \begin{center}
    \includegraphics[width=0.48\textwidth]{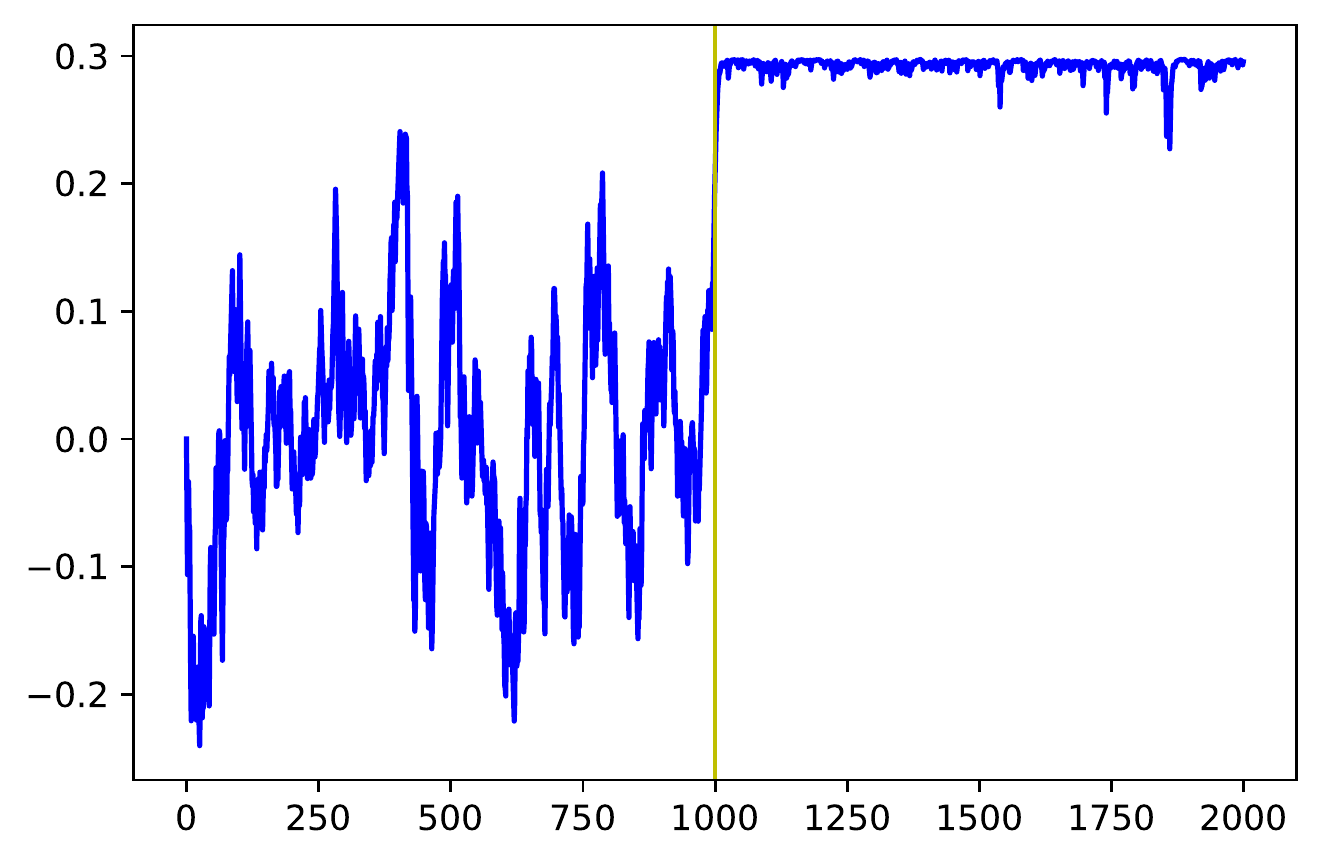}
    \includegraphics[width=0.48\textwidth]{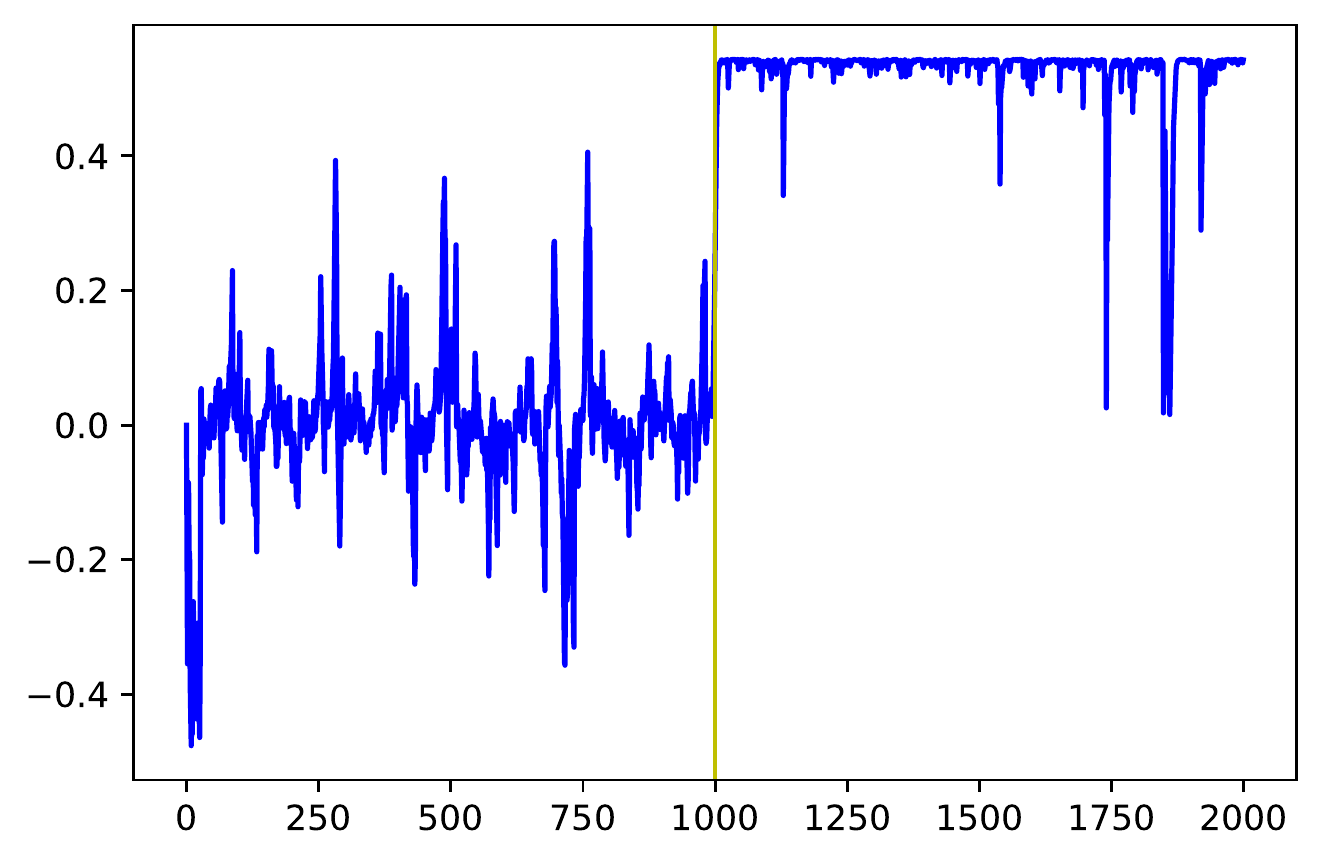}
  \end{center}
  \caption{The enhanced point predictions (medians of the enhanced predictive distributions).
    Left panel: $\epsilon\in\{-1,0,1\}$.
    Right panel: $\epsilon\in\{-2,0,2\}$.}
  \label{fig:prediction}
\end{figure}

Figure~\ref{fig:log-loss} provides useful and precise information,
but it is not very intuitive.
A cruder approach is to translate the probabilistic forecasts into point predictions.
Figure~\ref{fig:prediction} uses the medians of predictive distributions as point predictions.
In the case of the base algorithm,
the prediction is always 0 (the median of $N(0,1)$),
for the oracle algorithm it is 0 before the changepoint and 1 afterwards,
and for the enhanced algorithm the predictions are shown in the figure.
We can see that the right panel of Figure~\ref{fig:prediction} is a better approximation to the oracle predictions.

\begin{remark}
  To compute the point predictions shown in Figure~\ref{fig:prediction},
  we can use the representation of the betting function $b$ for the Simple Jumper in the form \eqref{eq:f} with
  \[
    \epsilon
    :=
    \frac{C_E - C_{-E}}{C_E + C_0 + C_{-E}}
    E,
  \]
  where $\{-E,0,E\}$ is the range of $\epsilon$
  (so that $E=1$ in the left-hand panels and $E=2$ in the right-hand ones).
  The indefinite integral of the betting function is
  \[
    B(v)
    =
    \int_0^v b(u) \dd u
    =
    \int_0^v
    \left(
      1 + \epsilon(u-0.5)
    \right)
    \ddd u
    =
    \left(
      1 - \frac{\epsilon}{2}
    \right)
    v
    +
    \frac{\epsilon}{2}
    v^2.
  \]
  Solving the quadratic equation $B(v)=0.5$ we get
  \begin{equation}\label{eq:v}
    v
    =
    \frac{\epsilon-2+\sqrt{\epsilon^2+4}}{2\epsilon}.
  \end{equation}
  Since the distribution function of the enhanced probability forecast is $B(F)$,
  where $F=N(0,1)$ is the distribution function of the original probability forecast,
  we obtain the median of the enhanced distribution as the $v$ quantile of $N(0,1)$,
  with $v$ defined by \eqref{eq:v}.
\end{remark}

\section{Conclusion}
\label{sec:conclusion}

This section briefly discusses possible directions of further research.

To understand better the potential of the new method,
further simulation studies and, more importantly, empirical studies are required.
In particular, this note uses only one proper loss function, namely the log loss function.
An interesting alternative is CRPS, or continuous ranked probability score \cite[Section 4.2]{Gneiting/Raftery:2007}.

Another direction is to improve the performance of test martingales and, therefore, enhanced prediction algorithms
in various model situations, similarly to \cite{\OCMXXXIII}.
The framework of Section~\ref{sec:simulation} is an example of such a model situation.

It is important to get rid of the assumption that the predictive distribution is continuous,
which we made in Section~\ref{sec:PIT}.
This is needed, e.g., to cover the case of classification.
This could be achieved by adapting the smoothing procedure \cite[(2.20)]{Vovk/etal:2005book},
which is standard in conformal prediction.

This note assumes that the observations $y_n$ are real numbers,
whereas the standard setting of machine learning is where the observations are pairs $(x_n,y_n)$.
Our method is applicable in this case as well if we assume that the $x_n$ are constant.
The cleanest approach, however, would be not to assume anything about the $x_n$
and use the game-theoretic foundations of probability \cite{Shafer/Vovk:2019}.
For the game-theoretic treatment of the probability integral transform,
see \cite[Theorem 2(b)]{Dawid/Vovk:1999}.

\subsection*{Acknowledgments}

This work has been supported by Amazon and Stena Line.


\begin{thebibliography}{10}
\bibitem{Dawid/Vovk:1999}
A.~Philip Dawid and Vladimir Vovk.
\newblock Prequential probability: {P}rinciples and properties.
\newblock {\em Bernoulli}, 5:125--162, 1999.

\bibitem{Gneiting/Raftery:2007}
Tilmann Gneiting and Adrian~E. Raftery.
\newblock Strictly proper scoring rules, prediction, and estimation.
\newblock {\em Journal of the American Statistical Association}, 102:359--378,
  2007.

\bibitem{Kolmogorov:1933}
Andrei~N. Kolmogorov.
\newblock {\em Grundbegriffe der Wahr\-schein\-lich\-keits\-rechnung}.
\newblock Springer, Berlin, 1933.
\newblock English translation: \emph{Foundations of the Theory of Probability}.
  Chelsea, New York, 1950.

\bibitem{Levy:1937}
Paul L\'evy.
\newblock {\em Th\'eorie de l'addition des variables al\'eatoires}.
\newblock Gauthier-Villars, Paris, 1937.
\newblock Second edition: 1954.

\bibitem{Rosenblatt:1952}
Murray Rosenblatt.
\newblock Remarks on a multivariate transformation.
\newblock {\em Annals of Mathematical Statistics}, 23:470--472, 1952.

\bibitem{Shafer/Vovk:2019}
Glenn Shafer and Vladimir Vovk.
\newblock {\em Game-Theoretic Foundations for Probability and Finance}.
\newblock Wiley, Hoboken, NJ, 2019.

\bibitem{Vovk:arXiv2012}
Vladimir Vovk.
\newblock Testing for concept shift online.
\newblock Technical Report
  \href{https://arxiv.org/abs/2012.14246}{arXiv:2012.14246 [cs.LG]},
  \href{https://arxiv.org}{arXiv.org} e-Print archive, December 2020.

\bibitem{Vovk:arXiv2104}
Vladimir Vovk.
\newblock Conformal testing in a binary model situation.
\newblock Technical Report
  \href{https://arxiv.org/abs/2104.01885}{arXiv:2104.01885 [cs.LG]},
  \href{https://arxiv.org}{arXiv.org} e-Print archive, April 2021.

\bibitem{Vovk:2021}
Vladimir Vovk.
\newblock Testing randomness online.
\newblock {\em Statistical Science}, 2021.
\newblock To appear, published online.

\bibitem{Vovk/etal:2005book}
Vladimir Vovk, Alex Gammerman, and Glenn Shafer.
\newblock {\em Algorithmic Learning in a Random World}.
\newblock Springer, New York, 2005.

\bibitem{Vovk/etal:arXiv2102}
Vladimir Vovk, Ivan Petej, Ilia Nouretdinov, Ernst Ahlberg, Lars Carlsson, and
  Alex Gammerman.
\newblock Retrain or not retrain: Conformal test martingales for change-point
  detection.
\newblock Technical Report
  \href{https://arxiv.org/abs/2102.10439}{arXiv:2102.10439 [cs.LG]},
  \href{https://arxiv.org}{arXiv.org} e-Print archive, February 2021.
\end{thebibliography}
\end{document}